\documentclass{bmvc2k}
\usepackage{multirow}
\usepackage{multicol}
\usepackage{tabularray}
\usepackage{tabularx}
\usepackage{adjustbox}

\title{X-PDNet: Accurate Joint Plane Instance Segmentation and Monocular Depth Estimation with Cross-Task Distillation and Boundary Correction}

\addauthor{Cao Dinh Duc}{duccd@hanyang.ac.kr}{1,3}
\addauthor{Jongwoo Lim}{jongwoo.lim@snu.ac.kr}{2}

\addinstitution{
 Dept of Computer Science\\
 Hanyang University\\
 Seoul, South Korea
}
\addinstitution{
 Dept of Mechanical Engineering\\
 Seoul National University\\
 Seoul, South Korea
}
\addinstitution{
 AI Research Lab, MAXST Co., Ltd\\
 Seoul, South Korea
}

\runninghead{Duc Cd, J Lim}{X-PDNet}

\begin{document}

\maketitle

\begin{abstract}
Segmentation of planar regions from a single RGB image is a particularly important task in the perception of complex scenes. To utilize both visual and geometric properties in images, recent approaches often formulate the problem as a joint estimation of planar instances and dense depth through feature fusion mechanisms and geometric constraint losses. Despite promising results, these methods do not consider cross-task feature distillation and perform poorly in boundary regions. To overcome these limitations, we propose X-PDNet, a framework for the multitask learning of plane instance segmentation and depth estimation with improvements in the following two aspects. Firstly, we construct the cross-task distillation design which promotes early information sharing between dual-tasks for specific task improvements. Secondly, we highlight the current limitations of using the ground truth boundary to develop boundary regression loss, and propose a novel method that exploits depth information to support precise boundary region segmentation. Finally, we manually annotate more than 3000 images from Stanford 2D-3D-Semantics dataset and make available for evaluation of plane instance segmentation. 
Through the experiments, our proposed methods prove the advantages, outperforming the baseline with large improvement margins in the quantitative results on the ScanNet and the Stanford 2D-3D-S dataset, demonstrating the effectiveness of our proposals. The code is available at: \url{https://github.com/caodinhduc/X-PDNet-official}.
\end{abstract}

\section{Introduction}
Piecewise planar regions frequently appear in man-made environments, especially in indoor scenes (wall, floor, furniture, etc.). The detection and segmentation of such piecewise planar surfaces in images has attracted much attention due to its wide range of applications.  
In the indoor environment, planar instance segmentation offers an essential representation for scene understanding~\cite{Tsai2011RealtimeIS}, augmented reality (AR) applications, robot navigation, and visual SLAM~\cite{Rambach2019SlamCraftDP}. In the outdoor scenes, ground and wall plane cues benefit 6-DoF object pose estimation, building reconstruction~\cite{Li2016ReconstructingBM}, and drivable surface detection in autonomous driving.
Recently, with the advancement of deep neural networks, the piecewise estimation of planar region can be reformulated to the plane instance segmentation task. Starting with PlaneNet~\cite{Liu2018PlaneNetPP} and PlaneRecover~\cite{Yang2018Recovering3P}, which make breakthroughs in using convolutional neural networks (CNNs) to segment planar or non-planar region instances. Next, PlaneRCNN~\cite{Liu2019PlaneRCNN3P} inherits Mask R-CNN~\cite{He2017MaskR} to segment plane instances with their plane parameters and segmentation masks. PlaneSegNet~\cite{Xie2021PlaneSegNetFA} builds upon Yolact++~\cite{Bolya2022YOLACTBR}, which was presented as the first real-time single-stage plane instance segmentation method in this field.
PlaneRecNet~\cite{Xie2021PlaneRecNetML} forms a multi-task learning framework by jointly training a single-stage plane instance segmentation network with depth estimation from a single RGB image. Unlike other existing approaches~\cite{Liu2018PlaneNetPP, Liu2019PlaneRCNN3P, Yang2018Recovering3P, Yu2019SingleImagePP}, PlaneRecNet concentrates on enforcing cross-task consistency by introducing multiple loss functions (geometric constraints) that cooperatively enhance the accuracy of plane instance segmentation and depth estimation. Despite achieving solid quantitative results on both tasks besides computational efficiency, PlaneRecNet still has several limitations that need to be improved. 1) Since the instance segmentation mask candidates are fused to hidden depth features through multiplication and concatenation computations, this design inherently limits the adaptive feature distillation capability between cross-tasks and further limits the performance of the plane instance segmentation while over-focusing on depth estimation. 2) Current single-stage plane instance segmentation methods do not explicitly utilize the boundary information of the ground truth masks, which results in imprecise predicted masks. Furthermore, because the ground truth plane masks are generated by RANSAC-based methods, it produces incorrect and coarse boundary ground truth instances. Hence, predicted masks optimized by traditional boundary regression loss not to be tightly aligned to the true boundaries. 
To address these issues, we propose X-PDNet (X indicates a cross design), a framework for joint plane instance segmentation and depth estimation, which is based upon PlaneRecNet~\cite{Xie2021PlaneRecNetML} with several major improvements. We introduce the cross-task distillation design, where distillation modules are dual-integrated between the aggregated depth feature layer and the feature mask layer of SOLO V2~\cite{Wang2020SOLOv2DA} network. In addition, we propose \textbf{Depth Guided Boundary Preserving Loss}, which alleviates the effect of incorrect ground truth masks by evaluating the gradient difference between the boundary ground truth and its neighbors at the pixel level.
Our main contributions can be summarized as follows:
\begin{itemize}
\vspace{-0.2cm}
\item Developing from PlaneRecNet~\cite{Xie2021PlaneRecNetML}, we design X-PDNet, a multi-task learning framework for joint plane instance segmentation and depth estimation, which allows the respective task decoder to adaptively distill the cross-supplementary information for the specific task optimization.
\vspace{-0.2cm}
\item We introduce a novel Depth Guided Boundary Preserving Loss, which combats noisy ground truth to produce more accurate segmentation at boundary regions.
\vspace{-0.2cm}
\item We contribute manual annotations of over 3000 images from the Stanford 2D-3D-Semantics dataset as a reliable evaluation set for plane instance and boundary segmentation. 
\vspace{-0.2cm}
\item Extensive experiments on the ScanNet and the 2D-3D-S datasets demonstrate the effectiveness of our method in both plane instance segmentation and depth estimation tasks by a large margin improvements over previous methods with no additional computational cost.
\end{itemize}
\section{Related Work}
\begin{figure}
\begin{center}
\bmvaHangBox{\fbox{\includegraphics[width=0.95\textwidth]{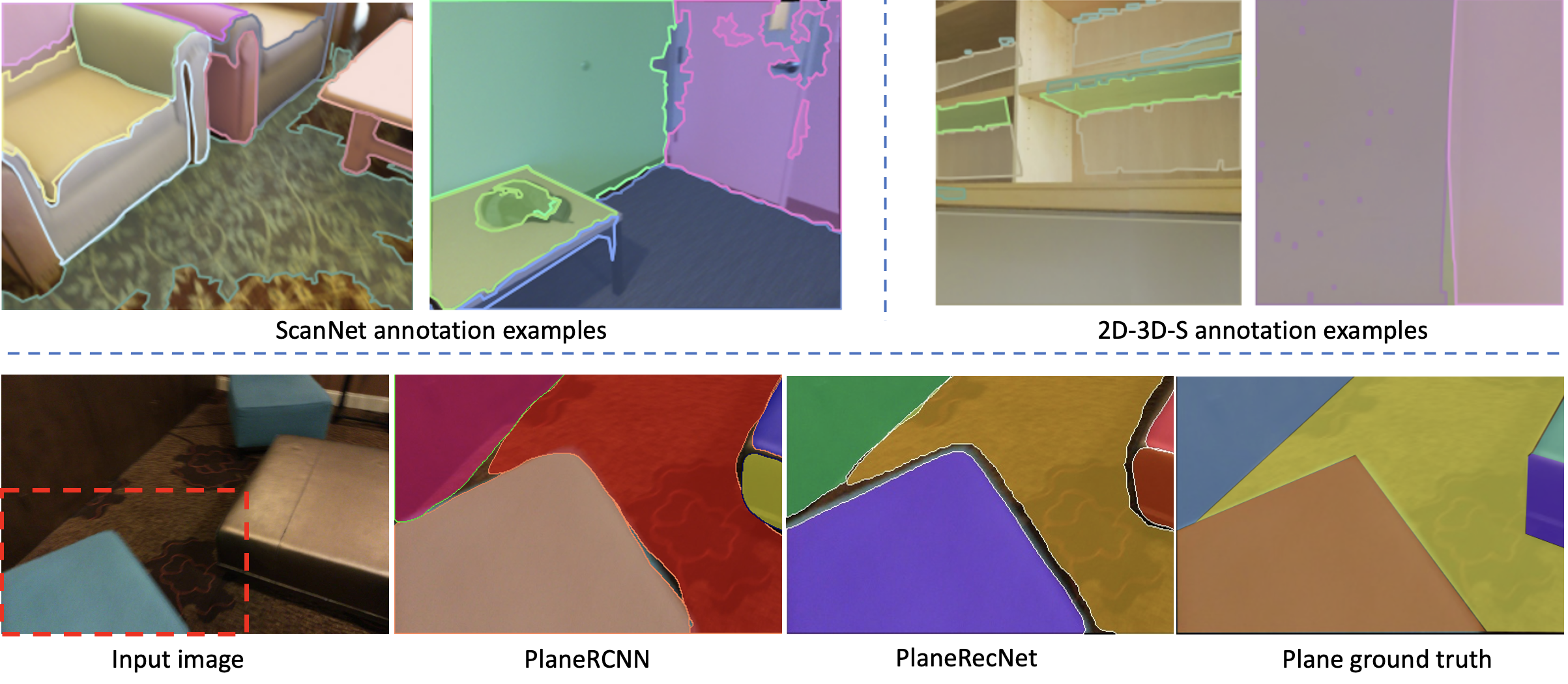}}}
\end{center}
\caption{The first row shows examples of incorrect instance ground truth from different datasets. The second row visualises the segmentation results produced by existing methods on the ScanNet~\cite{Dai2017ScanNetR3} dataset with poor quality predictions at the boundary regions of planes.}
\label{fig:coarse}
\end{figure}
{\bf Plane Instance Segmentation:} PlaneNet~\cite{Liu2018PlaneNetPP} is the first attempt employing a deep neural network to reconstruct piecewise planar regions from a single RGB image. It shares an encoder and provides three prediction branches: plane parameter estimation, plane segmentation, and non-planar depth map estimation. Later, in PlaneRecover~\cite{Yang2018Recovering3P}, Yang and Zhou indicate the obstacles to obtaining the ground truth of the plane annotation dataset. Then they present a novel plane structure-induced loss to train the plane segmentation and plane parameter estimation for outdoor scenes through an unsupervised learning approach. In spite of generating promising results, both PlaneNet and PlaneRecover require a fixed number of predicted planar regions, which severely restricts the generalization capabilities of the application to different scenarios. PlaneAE~\cite{Yu2019SingleImagePP} trains a CNN to map each pixel to an embedding space where pixels from the same plane instance have similar embeddings. It then groups embedding vectors into piecewise plane instances using its mean shift clustering algorithm. PlaneRCNN~\cite{Liu2019PlaneRCNN3P} proposes an effective plane segmentation branch built upon Mask R-CNN~\cite{He2017MaskR} and jointly refines the segmentation mask with their novel warping loss function. The method shows high localization ability and generalization across different domains but fails to achieve real-time execution. PlaneSegNet~\cite{Xie2021PlaneSegNetFA} introduces a fast single-stage instance segmentation method for high-resolution piece-wise planar regions, the approach adapts strongly at large-scale planar regions but misses depth estimation. Differently, PlaneRecNet~\cite{Xie2021PlaneRecNetML} designs a multi-task network for jointly studying plane instance segmentation and depth estimation and boosting the cross-task consistency by exploiting geometric constraints.
\vspace{0.2cm}
\newline{\bf Cross-Task Distillation Mechanism:} Related to our work are methods that facilitate feature sharing or distillation across tasks, inspired by the idea that each task could benefit from complementary information from the others. PAD-Net~\cite{Xu2018PADNetMG} uses an attention mechanism to distill information across multimodal features. MTI-Net~\cite{Vandenhende2020MTINetMT} extends PAD-Net with a multi-scale solution to better distill multimodal information.~\cite{Zhang2019PatternAffinitivePA} proposes to learn a single-task affinity matrix, then which is then combined to diffuse and refine the task-specific features.~\cite{Bruggemann2021ExploringRC} introduces ATRC to refine each task prediction by capturing cross-task contexts dependent on four relational context types.
\vspace{0.2cm}
\newline{\bf Boundary Preserving Loss for Segmentation:} Obtaining sharp boundaries is important for the high-quality instance segmentation task. Existing methods are solid in terms of plane localization but do not pay attention to the exploitation of boundary information. As a result, these models produce planes with coarse and imprecise contours that are typically illustrated by the overlaps or the gaps between two adjacent planes as shown in Fig.\ref{fig:coarse}. Observing in the segmentation field, enhancing segmentation accuracy in boundary regions has been studied in some existing methods~\cite{Zhang2021RefineMaskTH, Cheng2020BoundarypreservingMR, Kirillov2020PointRendIS, Takikawa2019GatedSCNNGS, Wang2022NoisyBL, Zhu2022SharpContourAC} but mostly developed for detection-then-segmentation methods. BMask R-CNN~\cite{Cheng2020BoundarypreservingMR} achieves a better result by combining the representation of object boundaries to guide mask prediction.  Gate-SCNN~\cite{Takikawa2019GatedSCNNGS} jointly supervised segmentation and boundary map prediction. ~\cite{Wang2022NoisyBL} introduces a boundary-preserving reweighting mechanism that forces the model to focus on boundary-relevant areas. BSOLO~\cite{Zhang2022BSOLOBO} designs a Hungarian algorithm based border loss to calculate the cost of matching between borders. While these methods show that they can lead to higher quality predicted masks, they still suffer from several limitations, including the high computational cost due to the additional branch for edge detection, the lack of ideal edge ground truth, and the unstable or low quality of predicted edges.
\begin{figure}
\begin{center}
\bmvaHangBox{\fbox{\includegraphics[width=0.95\textwidth]{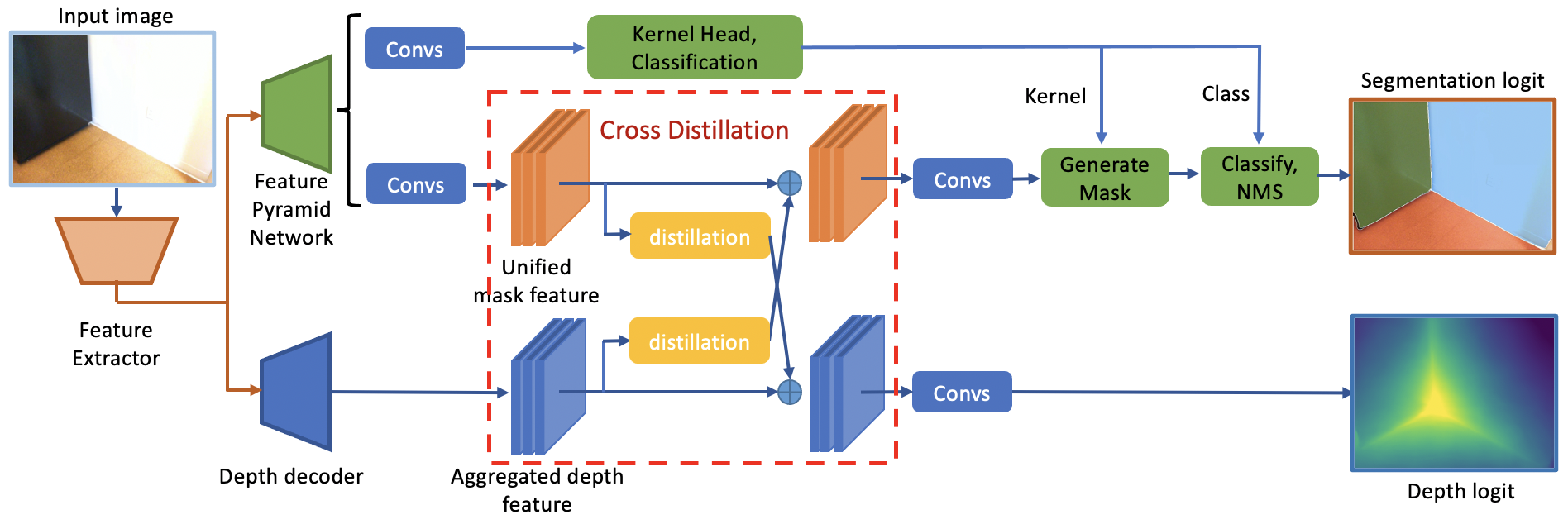}}}
\end{center}
\caption{The architecture of \textbf{X-PDNet}: The network consists of a shared backbone and two parallel branches for plane instance segmentation and monocular depth estimation. A couple of Cross Distillation Modules are integrated between the unified mask feature of the segmentation branch and aggregated feature of the depth decoder to facilitate early cross-task feature distillation. Detail of distillation module is described in the Fig.~\ref{fig:attention}.}
\label{fig:architecture}
\end{figure}
\section{Method}
\subsection{X-PDNet Overview}
Our proposed X-PDNet is built upon the PlaneRecNet~\cite{Xie2021PlaneRecNetML} with several major improvements to address the aforementioned problems. As described in Fig.~\ref{fig:architecture}, given a single color image as an input, our network consists of two branches with a shared backbone to predict a piece-wise planar segmentation $S_{pred}$ and a depth estimation $D_{pred}$ in parallel. A couple of distillation modules are dual-integrated between the aggregated feature of the depth decoder and the mask feature of the segmentation branch to distill cross-task complementary signals.
\subsection{Cross Distillation Design}
\begin{figure}
\begin{center}
\bmvaHangBox{\fbox{\includegraphics[width=0.8\textwidth]{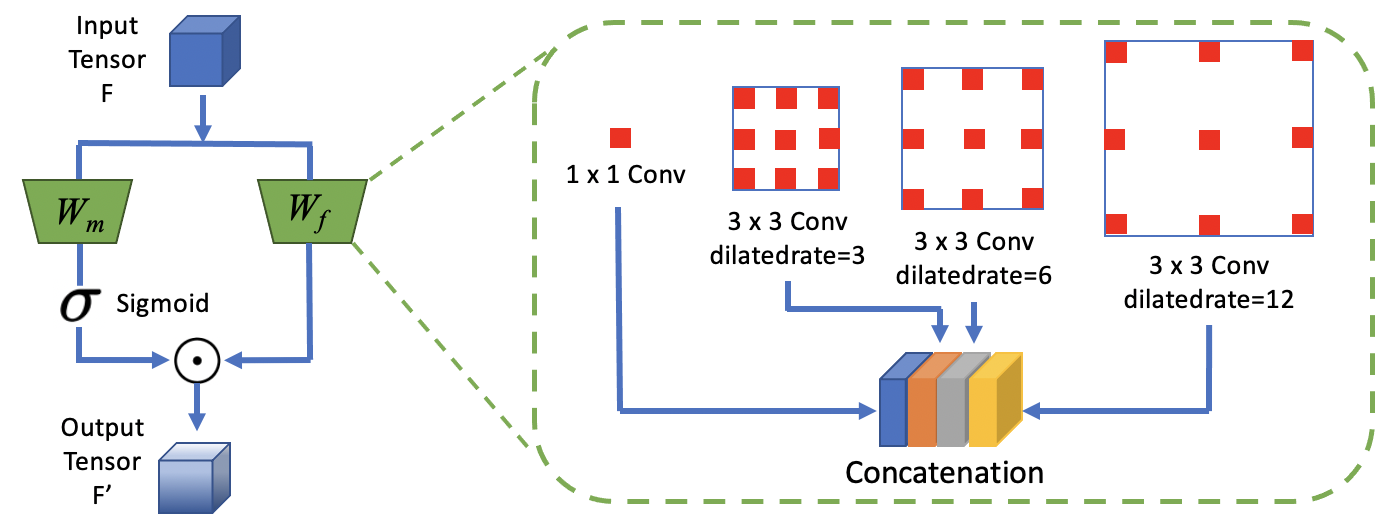}}}
\end{center}
\caption{Illustration of the cross-task distillation module, which involves a feature branch (convolution layers with different dilated rates (green box)) and mask branch (a convolution layer with a sigmoid function to construct an attention mask), and the output feature is combined by the element-wise multiplication.}
\label{fig:attention}
\end{figure}
Attention or distillation mechanisms~\cite{Xu2018PADNetMG, Vandenhende2020MTINetMT, Bruggemann2021ExploringRC, Zhang2019PatternAffinitivePA, Lopes2022CrosstaskAM} have been commonly used to facilitate cross-task optimization in multi-task learning for a long time, this builds on the intuition that each decoder could learn from the complementary signal of another branch. Moreover, since the cross-task feature is not always beneficial for the primary task, the distillation module can act as a filter to select only useful information from the other tasks.
Considering the baseline method~\cite{Xie2021PlaneRecNetML}, which is based on SOLO V2~\cite{Wang2020SOLOv2DA} for the plane instance segmentation branch, where the mask candidates are fused into the depth branch through the Plane Prior Attention module, we argue that fusing plane-predicted masks into depth aggregated features imposes the model optimize for the depth estimation task but may affect the segmentation accuracy due to the depth backpropagation gradient through the plane instance mask candidates. To facilitate the early cross-task information distillation, in our work, we introduce a lightweight but efficient cross-distillation design to guide the message passing between the aggregated feature maps generated by the depth decoder and feature mask of the segmentation branch as illustration in Fig.~\ref{fig:architecture}. We leverage the idea presented in PAD-NET~\cite{Xu2018PADNetMG} with a reformulation to help the model adapt robustly with multiple scale plane instances as reported in~\cite{Xie2021PlaneSegNetFA}. Given the context that we want to pass the message from the secondary task to facilitate the primary task. As visualization in Fig.~\ref{fig:attention}, firstly, an attention map (the output of sigmoid function) $A$ is generated from the secondary task feature $F$ as follows:
\begin{equation}
  A \leftarrow \sigma(W_m \otimes F),
\end{equation}
Where $W_m$ is the 2-D convolution parameters and $\sigma$ is a sigmoid function to normalize the attention map. Then the message passed from the secondary task $F$ is controlled by the attention map $A$ as described:
\begin{equation}
  F' \leftarrow A \odot (W_f \otimes F).
\label{equa:2}
\end{equation}
In the equation~\ref{equa:2}, inspired by the ASPP design~\cite{Chen2017RethinkingAC}, we extract the cross-task feature $F$ by a set of convolution layers ($W_f$) with different dilation rates ([1, 3, 6, 12] in our experiments) to enlarge the spatial scale of cross-task contexts, then stack the outputs together as depiction in the Fig.~\ref{fig:attention}, $\odot$ and $\otimes$ denote the element-wise multiplication and convolution operation. Finally, the passed message $F'$ is merged into the primary task for specific task optimization as shown in Fig.~\ref{fig:architecture}. Through the experiments in Tab.~\ref{tab:ablation}, we demonstrate the effectiveness of our design as well as the importance of receptive field expansion in the cross-task feature distillation by significantly improving from not only the segmentation but also the depth estimation.
\subsection{Depth Guided Boundary Preserving Loss (\textbf{DGBPL})}
\begin{figure}
\begin{center}
\bmvaHangBox{\fbox{\includegraphics[width=0.95\textwidth]{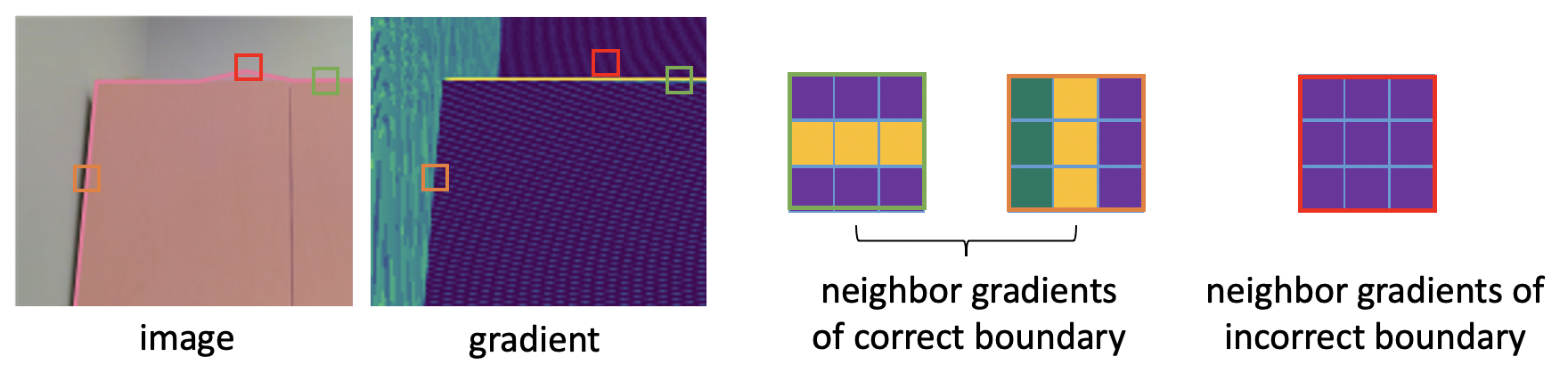}}}
\end{center}
\caption{Visualization of gradient based analysis. With a boundary ground truth point (center of each window), we consider its depth gradient and that of its neighbors within a window, yellow elements indicate boundary points.}
\label{fig:dgbpl}
\end{figure}
The traditional approach (\textbf{Vanilla}) to the problem of poor performance at boundary regions is boundary regression loss, which obtains the ground truth and predicted boundaries by the edge detectors (Sobel, Laplacian), then uses Mean Square Error to teach the predicted masks to align with the true boundaries. However, since the quality of the GT masks is poor (Fig.~\ref{fig:coarse} first row), it forces the segmentation method trained with the vanilla method produce segmentation masks that is far from ideal as example in Fig.~\ref{fig:example_1}. To alleviate this limitation, we evaluate the confidence of the ground truth boundaries at pixel level by measuring the difference between its depth gradient with that of its neighbors. We observe that the depth gradient fluctuates slightly over the plane area but changes suddenly at the occluded areas or junctions between adjacent planes (Fig.~\ref{fig:dgbpl}). To exploit this constraint, we first construct the gradient mask $G_{gt}$ from the ground truth depth $D_{gt}$ using Sobel-Filter:
\begin{equation}
    G_{gt} = abs(G_x) + abs(G_y) \;\;\; \mathrm{with} \;\; G_x = Sobel_x(D_{gt}),\; G_y = Sobel_y(D_{gt}).
\label{gradient}
\end{equation}
As presented in equation.~\ref{gradient}, we formulate this mask as a combination of absolute gradients following the x and y directions. Next, for each pair ground truth mask $y^{gt}_m \in (H/4\times W/4)$  and predicted mask $y^{pr}_m\in (H/4\times W/4)$ generated by the plane instance segmentation branch, we follow the traditional edge detection method (Laplacian operation) to obtain the ground truth boundary (${y^{gt}_b}$) and the predicted boundary (${y^{pr}_b}$), respectively. For each boundary point in (${y^{gt}_b}$), we consider the local gradient variation by obtaining the corresponding gradient values within the window (3x3 with the target point at the center in our experiment), then measure the standard deviation of these points. As visualization in Fig.~\ref{fig:dgbpl}, we expect the standard deviation (std) computed from the correct GT boundary (green and orange boxes) will be higher than that computed from the incorrect GT boundary (red box). We then normalize these std values to estimate the weights ($W$) before using them to reweight the boundary regression loss (MSE(${y^{gt}_b} * W$, ${y^{pr}_b * W}$) ) at the pixel level to guide the this loss to focus on the correct contour while reducing the impact of noise ground truth boundaries.
DGBPL mitigates the impact of an imperfect plane GT mask on normal boundary regression loss, as demonstrated by better results on a manually annotated dataset while maintaining the same training set. See section.~\ref{dgbpl-eval} for a detailed evaluation.
\section{Experiments Setup}
\subsection{Datasets and metrics}
To measure the performance of our proposals, we conduct experiments on two public datasets: ScanNet with annotation provided by~\cite{Liu2019PlaneRCNN3P} and 2D-3D-S with annotation provided by~\cite{Xie2021PlaneSegNetFA}. For the 2D-3D-S dataset, we additionally test the networks on our manually annotated evaluation set to figure out clearly the effectiveness of the proposed loss function. For the quantitative metrics in plane instance segmentation, we use Average Precision for both masks $(AP_m)$ and bounding boxes $(AP_b)$ at different NMS thresholds (overall, 50, and 75). In terms of depth estimation evaluation, the metrics include Absolute Relative Error $(rel)$, Log 10 error $(log_{10})$, linear Root Mean Square Error $(RMS)$, and accuracy under the thresholds $(\delta_1, \delta_2, \delta_3)$.
\subsection{Implementation details}
Similar to PlaneRecNet~\cite{Xie2021PlaneRecNetML}, our proposed X-PDNet is implemented using the Pytorch~\cite{Paszke2019PyTorchAI} framework. It adopts ResNet101~\cite{He2016DeepRL} with deformable convolution~\cite{Zhu2019DeformableCV} as the backbone network. We use Adam optimizer~\cite{Kingma2015AdamAM} and a batch size of 8 images for model training. For a fair comparison, we keep the loss functions, loss weights, and training strategies from the baseline~\cite{Xie2021PlaneRecNetML}, To be more specific, losses include:
\begin{equation}
    L = L_{Focal} + L_{Dice} + L_{RMSE} + L_{constraints} + DGBPL
\label{loss}
\end{equation}
Where focal and dice losses are for the segmentation task, RMSE is for the optimization of the depth estimation, and geometric constraint losses.
Our model is trained for 10 epochs on ScanNet and 15 epochs on 2D-3D-S with the plane annotation given by~\cite{Liu2019PlaneRCNN3P} and ~\cite{Xie2021PlaneSegNetFA}, respectively. In both datasets, training data is augmented with random photometric distortion, horizontal and vertical flipping, and Gaussian noise. All training sessions are conducted on an NVIDIA RTX A5000 GPU device.
\section{Experiments}
\subsection{Comparison with existing methods}
\begin{figure}[t]
\begin{center}
\bmvaHangBox{\fbox{\includegraphics[width=0.95\textwidth]{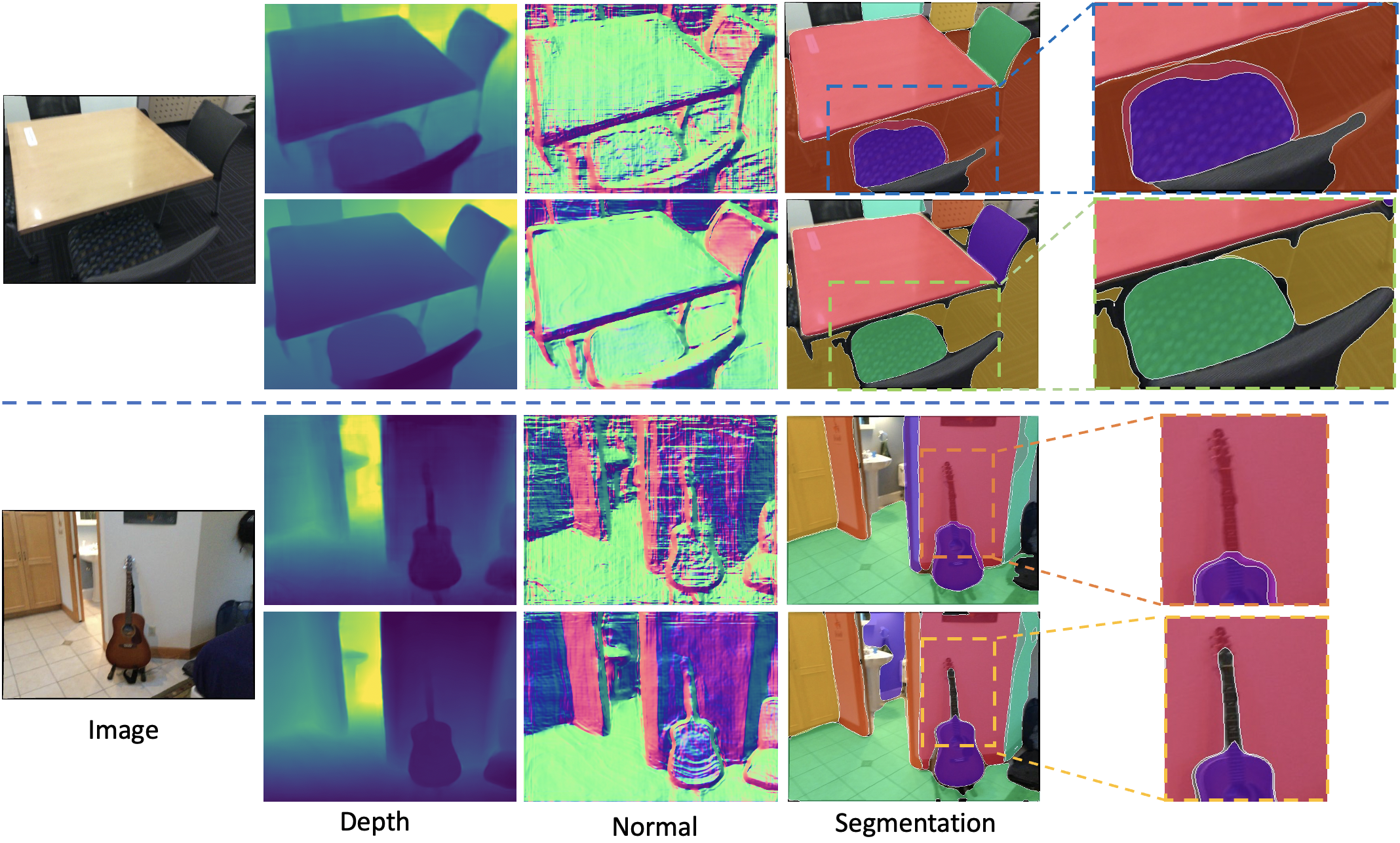}}}
\end{center}
\caption{Qualitative comparison between X-PDNet and the baseline (PlaneRecNet) on images from the ScanNet dataset. It contains two examples, for each one, the first row is the output of \textbf{PlaneRecNet}, while the second row is generated by \textbf{X-PDNet}, (normal is recovered from the predicted depth). The obvious difference can be seen in the rectangle boxes. In the chair area of the first example, with cross-task feature distillation, X-PDNet is able to distinguish the chair surface from the floor, even though the RGB feature in this area is quite similar. In terms of depth, ours is better with the perception of visual information from the segmentation branch, resulting in a smooth normal vector converted from depth prediction in each plane area. The same improvement is observed in the second example (guitar surface).}
\label{fig:example}
\end{figure}

\begin{table}[h]
\begin{adjustbox}{width=1.0\columnwidth}
\begin{tabular}{|l|c|cccccc|cccccc|}
\hline
\multirow{2}{*}{Methods} &
\multirow{2}{*}{Dataset} &
\multicolumn{6}{c|}{Segmentation Metrics} &
\multicolumn{6}{c|}{Depth Metrics}\\
&& $AP_m$ & $AP_m^{50}$ & $AP_m^{75}$ & $AP_b$  & $AP_b^{50}$ & $AP_b^{75}$ & $rel$ $\downarrow$ & $log_{10}$ $\downarrow$ & $RMS$ $\downarrow$ & $\delta 1$ & $\delta 2$ & $\delta 3$ \\
\hline\hline
PlaneAE~\cite{Yu2019SingleImagePP} & ScanNet & 5.92 & 14.72 & 4.00 & 7.86 & 17.83 & 6.25 & 0.111 & 0.049 & 0.409 & 0.864 & 0.967 & 0.991\\
PlaneRCNN~\cite{Liu2019PlaneRCNN3P} & ScanNet & 14.23 & 28.23 & 12.88 & 17.51 & 33.00 & 16.00 & 0.124 & 0.050 & 0.265 & 0.865 & 0.972 & 0.994\\
PlaneRecNet~\cite{Xie2021PlaneRecNetML} & ScanNet & 16.61 & 31.59 & 15.56 & 21.05 & 36.45 & 20.29 & 0.076 & 0.032 & 0.180 & 0.950 & 0.992 & 0.998\\
\textbf{X-PDNet} & ScanNet & \textbf{17.62} & \textbf{33.05} & \textbf{16.60} & \textbf{22.23} & \textbf{37.53} & \textbf{21.91} & \textbf{0.069} & \textbf{0.029} & \textbf{0.175} & \textbf{0.955} & \textbf{0.993} & \textbf{0.999}\\
\hline \hline
PlaneRecNet~\cite{Xie2021PlaneRecNetML} & 2D-3D-S & 24.10 & 38.99 & 24.39 & 27.13 & 41.14 & 27.23 & 0.062 & 0.027 & \textbf{0.294} & \textbf{0.966} & 0.990 & \textbf{0.996}\\
\textbf{X-PDNet} & 2D-3D-S & \textbf{25.20} & \textbf{39.63} & \textbf{25.79} & \textbf{28.62} & \textbf{41.80} & \textbf{29.15} & \textbf{0.061} & \textbf{0.026} & \textbf{0.294} & \textbf{0.966} & \textbf{0.991} & \textbf{0.996}\\
\hline
\end{tabular}
\end{adjustbox}
\vspace{-0.2cm}
\caption{Evaluation of plane instance segmentation and depth estimation on \textbf{ScanNet} and \textbf{2D-3D-S} datasets. \textbf{X-PDNet} outperforms existing methods in most metrics.}
\label{tab:dgbpl}
\end{table}
This section is to illustrate the effectiveness of Cross Distillation Design, proved by remarkable improvements over the existing approaches on evaluation datasets. We first evaluate our proposed model on the ScanNet dataset which is the most popular dataset in plane instance segmentation with annotation generated by~\cite{Liu2019PlaneRCNN3P}. We utilize the same data setup with~\cite{Liu2019PlaneRCNN3P} and~\cite{Xie2021PlaneRecNetML}, which contains 100,000 training and 5,000 test images. Next, we further conduct the experiments on the 2D-3D-S dataset, which includes 60,000 training images and 5,000 test images. As the quantitative results are shown in Tab.~\ref{tab:dgbpl}, for the ScanNet dataset, X-PDNet outperforms the existing methods by a large margin in both task plane instance segmentation and depth estimation. Furthermore, for the 2D-3D-S dataset, there is still a large improvement in the segmentation performance of X-PDNet compared to the baseline, While the performance in terms of depth metrics increases slightly. Fig.~\ref{fig:example} shows the qualitative improvements of \textbf{X-PDNet} compared to the baseline (\textbf{PlaneRecNet}). With Cross Distillation Design, the segmentation estimator has better geometric understanding to predict more accurate plane masks, especially in occluded areas or areas where RGB information is ambiguous. Meanwhile, perception of visual information allows the depth logit to be smoother in planar areas, resulting in noise reduction.
\newpage
\subsection{Evaluation of Depth Guided Boundary Preserving Loss}
\label{dgbpl-eval}
\begin{figure}[h]
\begin{center}
\bmvaHangBox{\fbox{\includegraphics[width=0.95\textwidth]{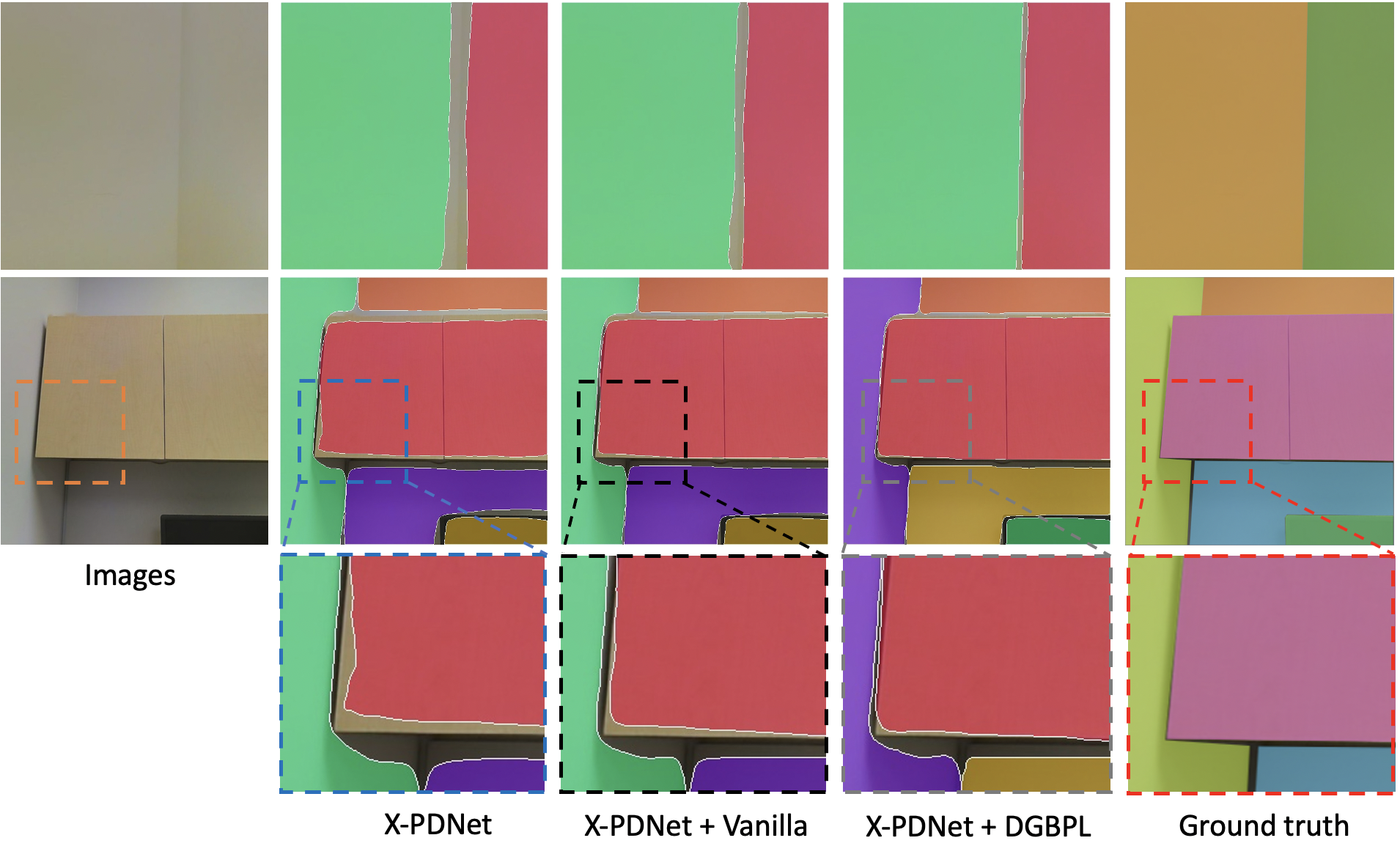}}}
\end{center}
\caption{Effect of Depth Guided Boundary Preserving Loss (\textbf{DGBPL}) on the segmentation results on 2D-3D-S examples compared to traditional regression boundary loss. With (\textbf{DGBPL}), X-PDNet performs impressively at boundary related regions. Focus on rectangle boxes for clear difference.}
\label{fig:example_1}
\end{figure}

Because the annotation of plane instances at boundary regions of existing datasets is under-qualified to verify the contribution of Depth Guided Boundary Preserving Loss. We provide a manually annotated label on the 2D-3D-S dataset. It is separate from the training and test set provided by \cite{Xie2021PlaneSegNetFA}. To measure how the depth gradient is beneficial in guiding the boundary preserving loss. We compare the segmentation performance of X-PDNet adding \textbf{DGBPL}, with and without vanilla boundary regression loss. In addition to segmentation metrics, we use the Boundary IoU $((y^{gt}_b\cap y^{pr}_b)/(y^{gt}_b\cup y^{pr}_b))$ measurement to claim that our proposed technique is beneficial for boundary region prediction. The details of the comparison given in Tab.~\ref{tab:boundary-evaluation} show that \textbf{DGBPL} outperforms the vanilla regression loss in the manually annotated, while being competitive in the original evaluation set. When guided by the depth gradient, \textbf{X-PDNet} produces the correct predictions in boundary regions, as shown by the reduced overlaps and narrow apertures between two adjacent planes as examples in Fig.~\ref{fig:example_1}. Examples of the original planar ground truth and after manual correction can be found in the supplementary document. 
\begin{table}[h]
\begin{adjustbox}{width=\columnwidth,center}
\begin{tabular}{|l|c|c|cccccc|}
\hline
\multirow{2}{*}{Methods} &
\multirow{2}{*}{Eval set} &
\multirow{2}{*}{Boundary IoU} &
\multicolumn{6}{c|}{Segmentation Metrics}\\
&&& $AP_m$ & $AP_m^{50}$ & $AP_m^{75}$ & $AP_b$  & $AP_b^{50}$ & $AP_b^{75}$ \\
\hline\hline
X-PDNet &Provided by~\cite{Xie2021PlaneSegNetFA}&-& 25.20 & 39.63 & 25.79 & 28.62 & 41.80 & 29.15 \\
X-PDNet+Vanilla & Provided by~\cite{Xie2021PlaneSegNetFA} &-& \textbf{26.49} & 41.61 & \textbf{27.09} & \textbf{30.23} & 44.18 & \textbf{30.7}\\
\textbf{X-PDNet+DGBPL} &Provided by~\cite{Xie2021PlaneSegNetFA}&- & 25.86 & \textbf{41.79} & 26.34 & 29.94 & \textbf{45.55} & 29.98\\
\hline \hline
X-PDNet & Manually annotated & 13.36& 24.09 & 36.84 & 25.08 & 25.80 & 37.08 & 26.72\\
X-PDNet+Vanilla & Manually annotated & 14.82 & 25.27 & 38.24 & 26.59 & 27.08 & 38.93 & \textbf{27.77}\\
\textbf{X-PDNet+DGBPL} & Manually annotated &\textbf{16.68}& \textbf{26.12} & \textbf{39.47} & \textbf{26.68} & \textbf{28.18} & \textbf{40.86} & 27.46\\
\hline
\end{tabular}
\end{adjustbox}
\begin{center}
\caption{Evaluation of segmentation results on \textbf{2D-3D-S} annotation provided by~\cite{Xie2021PlaneSegNetFA} and human labelling evaluation datasets.}
\label{tab:boundary-evaluation}
\end{center}
\end{table}
\vspace{-1.5cm}
\section{Ablation Study}
\label{Ablation}
Since our distillation module is based on the attention-guided message passing mechanism introduced in PAD-Net~\cite{Xu2018PADNetMG}. In this section, we analyze how our proposed modification affects the performance of joint instance segmentation and depth estimation. Specifically, we train the baseline \textbf{PlaneRecNet} (Plane Prior Attention), with no attention, cross design with attention module presented in PAD-Net~\cite{Xu2018PADNetMG}, and our (\textbf{X-PDNet}). The comparison shown in Tab.~\ref{tab:ablation} demonstrates the effectiveness of the cross-task distillation module (Fig.~\ref{fig:attention}) through the quantitative improvements in both depth and segmentation metrics. Refer the supplementary material for detail architecture of each design.
\begin{table}[h]
\begin{adjustbox}{width=\columnwidth,center}
\begin{tabular}{|l|cccccc|cccccc|}
\hline
\multirow{2}{*}{Attention/ distillation} &
\multicolumn{6}{c|}{Segmentation Metrics} &
\multicolumn{6}{c|}{Depth Metrics}\\
& $AP_m$ & $AP_m^{50}$ & $AP_m^{75}$ & $AP_b$  & $AP_b^{50}$ & $AP_b^{75}$ & $rel$ $\downarrow$ & $log_{10}$ $\downarrow$ & $RMS$ $\downarrow$ & $\delta 1$ & $\delta 2$ & $\delta 3$ \\
\hline\hline
No Attention or distillation  & 16.05 & 30.38 & 14.99 & 20.82 & 35.77 & 19.86 & 0.078 & 0.033 & 0.183 & 0.950 & 0.992 & 0.997\\
Plane Prior Attention~\cite{Xie2021PlaneRecNetML} & 16.61 & 31.59 & 15.56 & 21.05 & 36.45 & 20.29 & 0.076 & 0.032 & 0.180 & 0.950 & 0.992 & 0.998\\
PAD-Net~\cite{Xu2018PADNetMG}  & 17.41 & 32.54 & 16.48 & 22.11 & 37.24 & \textbf{21.96} & 0.071 & 0.031 & 0.176 & \textbf{0.955} & 0.992 & 0.998\\
\textbf{Ours} & \textbf{17.62} & \textbf{33.05} & \textbf{16.60} & \textbf{22.23} & \textbf{37.53} & 21.91 & \textbf{0.069} & \textbf{0.029} & \textbf{0.175} & \textbf{0.955} & \textbf{0.993} & \textbf{0.999}\\
\hline
\end{tabular}
\end{adjustbox}
\begin{center}
\caption{Ablation study of the performance of the network with different selection of attention or distillation designs on \textbf{ScanNet} dataset. \textbf{Ours} performs better in both tasks.}
\label{tab:ablation}
\end{center}
\end{table}
\section{Conclusion}
In this paper, we present two techniques to achieve precise joint learning of plane instance segmentation and depth estimation. We formulate a cross-task distillation design and explicitly exploit the depth information support for accurate segmentation at boundary related regions. Through extensive experiments, we demonstrate the effectiveness of our proposals by a considerable improvement in both tasks compared to the baselines.
\section{Acknowledgments}
This work was partially supported by the Technology Innovation Program (No. 20018110, "Development of a wireless tele-operable relief robot for detecting searching and responding in narrow space") funded by the Ministry of Trade, Industry \& Energy (MOTIE, Korea) and National Research Foundation of Korea (NRF) grant funded by the Korea government (MSIT) (NRF-2021R1A2C2010245).
\newpage
\bibliography{egbib}
\end{document}